\documentclass[runningheads]{llncs}

\usepackage{eccv}

\usepackage{amsmath}

\usepackage[linesnumbered,ruled,vlined]{algorithm2e}

\usepackage{xcolor}
\definecolor{mydarkblue}{RGB}{0,0,160} 

\SetCommentSty{mycommfont}

\newcommand{\under}[1]{\underline{\smash{#1}}}

\usepackage{eccvabbrv}

\usepackage{graphicx}
\usepackage{booktabs}

\usepackage[accsupp]{axessibility}  

\usepackage{hyperref}

\usepackage{orcidlink}
\usepackage{color}
\usepackage{enumitem}

\begin{document}

\title{Evolving Interpretable Visual Classifiers with Large Language Models} 

\titlerunning{Evolving Interpretable Visual Classifiers with LLMs}

\author{Mia Chiquier \and Utkarsh Mall \and Carl Vondrick}
\institute{Columbia University\\
\email{\{mia.chiquier,utkarshm,vondrick\}@cs.columbia.edu}}
\maketitle

\begin{abstract}
Multimodal pre-trained models, such as CLIP, are popular for zero-shot classification due to their open-vocabulary flexibility and high performance. However, vision-language models, which compute similarity scores between images and class labels, are largely black-box, with limited interpretability, risk for bias, and inability to discover new visual concepts not written down. Moreover, in practical settings, the vocabulary for class names and attributes of specialized concepts will not be known, preventing these methods from performing well on images uncommon in large-scale vision-language datasets. To address these limitations, we present a novel method that discovers interpretable yet discriminative sets of attributes for visual recognition. We introduce an evolutionary search algorithm that uses a large language model and its in-context learning abilities to iteratively mutate a concept bottleneck of attributes for classification. Our method produces state-of-the-art, interpretable fine-grained classifiers. We outperform the latest baselines by $18.4\%$ on five fine-grained iNaturalist datasets and by $22.2\%$ on two KikiBouba datasets, despite the baselines having access to privileged information about class names.

\end{abstract}

\section{Introduction}
\label{sec:intro}
 

Multimodal foundation models like CLIP \cite{radford2021learning} obtain excellent performance on many visual recognition tasks due to their flexibility to represent open-vocabulary classes. These models have the potential to impact many scientific applications, where computer vision systems could automate recognition in specialized domains. However, since foundation models are neural networks, they are largely black-box and we therefore have no means to explain or audit the predictions they produce, limiting their trust. Moreover, given that foundation models are trained on large corpora of web-scraped data \cite{radford2021learning, schuhmann2022laion}, they are not optimized to represent rare and fine-grained concepts, such as images from various scientific fields, which are infrequently described on the internet.

The computer vision community has been building interpretable models by integrating language, where classifiers are constructed with a bottleneck of sparse or discrete attributes \cite{ferrari2007learning,farhadi2009describing,menon2022visual,koh20concept,roth2023waffling}. Language-based approaches have the benefit of being interpretable. However, they rely on attributes that are either hand-designed, requiring expert knowledge, or extracted from external sources, such as large language models (LLMs). Since the attributes are not learned, they often obtain poor performance on specialized classes infrequently discussed in web-scale training sets.

\begin{figure}[t!]
  \centering
  \includegraphics[width=1.0\textwidth]{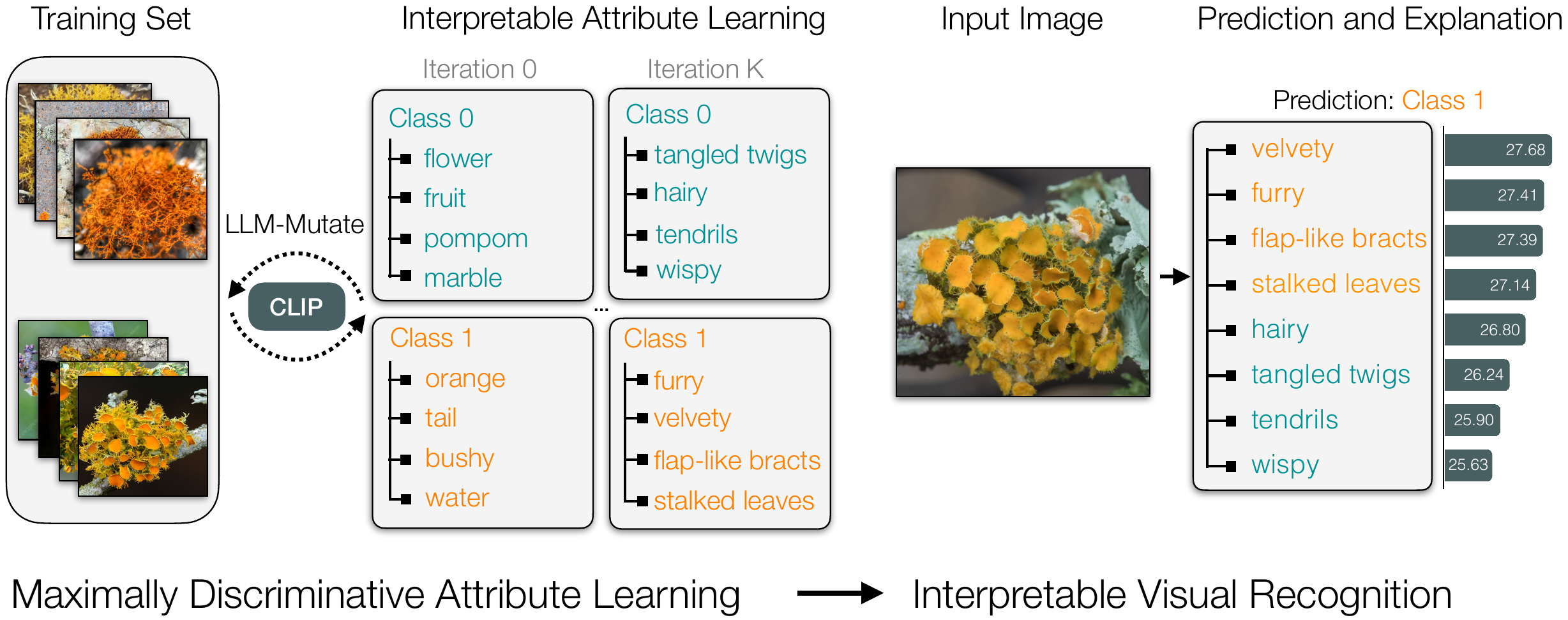}
  \caption{\textbf{Learning Interpretable Classifiers}. Can we find text attributes for a concept by looking at the images without their class names? LLM-Mutate is a framework that learns sets of maximally discriminative visual attributes per class without access to class names or any form of prior knowledge. }
  \label{fig:teasor}
\end{figure}

In this paper, we propose a framework to learn interpretable visual representations from images. Our approach can discover discrete, discriminative attributes from purely visual training sets of specialized concepts, and even unwritten concepts that do not appear on the internet. Solving this problem has typically been challenging because the loss is not differentiable with respect to discrete attributes, limiting the application of modern deep learning methods. Gradient-free methods, such as evolutionary search \cite{holland1992adaptation}, are able to optimize over discrete search spaces, but they have not scaled to large problems in the past.

Our approach integrates large language models with evolutionary algorithms in order to learn discriminative and interpretable attributes for visual recognition. Evolutionary algorithms work by maintaining a set of candidates, randomly mutating the candidates, and discarding the poorly performing solutions according to an objective function. The mutation step is the bottleneck for large-scale problems, such as in object recognition, because the search space is large and the lack of gradient information means the mutations are not guaranteed to drive the optimization towards rapid convergence.

We overcome these bottlenecks by replacing the mutation step with a large language model instead, whose in-context learning abilities are able to find patterns in-between candidates and predict strong mutations that reduce the objective. We evaluate our method on images from specialized scientific domain that have been infrequently discussed on the internet due to the nature of their specificity, with the iNaturalist dataset \cite{van2021benchmarking}. We chose five families of plants and animals within iNaturalist, each containing between five and six species, and evaluated the fine-grained classification performance on each family. We outperform all baselines, on average by $18.4\%$ per family dataset. We also evaluated our method of imaginary concepts that do not exist, and consequently have been hardly discussed in training sets before. Following the KikiBouba  experiments \cite{alper2024kiki} where people associate imaginary objects with non-existent words, we learn interpretable attributes that achieve strong discriminative performance, outperforming baselines by an average of $22.2\%$.

Our primary contribution is a framework for learning interpretable visual recognition systems from image data. We propose to tightly integrate evolutionary search with large language models, allowing us to efficiently learn discrete, discriminative attributes that are interpretable. The remainder of the paper will discuss the related work in this area, describe the method, and present experimental results on multiple datasets. 
Our code, data, and models are available at \href{https://llm-mutate.cs.columbia.edu}{https://llm-mutate.cs.columbia.edu}.

\section{Related Work}

We briefly review related work in interpretable methods, vision-language models and evolutionary search. 

\textbf{Concept Bottlenecks.} A common approach for interpretability is to create a bottleneck in the classifier, where interpretable attributes are first predicted before classifying the object category \cite{ferrari2007learning,farhadi2009describing,koh20concept, huang2016part,zhou2018interpretable,tang2020revisiting}.
Concept bottleneck models have been comprehensively studied in the domain of zero-shot learning~\cite{lampert-13,frome-13,akata-15,romera-15,kodirov-17}.
However, all these methods require a significant amount of annotations for the intermediate attributes to perform well while being interpretable.
Our method instead finds interpretable intermediate concepts by learning from visual data.  

\textbf{Post-hoc Interpretabililty.} Several post-hoc methods have been proposed to interpret deep models as well.
Methods such as GradCAM~\cite{selvaraju2017grad} rely on the activation maps to provide explanations for the classification on an image~\cite{chefer2021generic,fong2019understanding,zhou2016learning,petsiuk2018rise,shitole2021one,Simonyan2013DeepIC}.
Generating counterfactual examples is another way to interpret a model~\cite{goyal2019counterfactual,prabhu2024lance,vandenhende2022making,wang2020scout}.
Other work has looked at sample importance for explanations~\cite{yeh2018representer, Tsai2023SampleBE, Sui2021RepresenterPS, Pruthi2020EstimatingTD, koh2017understanding, Silva2020CrossLossIF, Guo2020FastIFSI}.
All these methods look at explanations from the perspective of a single image. 
In contrast to these approaches, several methods aim at understanding the behaviors of individual neurons in the trained model~\cite{gandelsman2023interpreting,dravid2023rosetta,zeiler2014visualizing,Frankle2018TheLT,Nguyen2016SynthesizingTP}. 
One major limitation of post-hoc interpretability methods they can only explain the receptive field of the model, which requires some interpretation on the user's part.
Concept or attribute bottleneck models on the other hand can justify the classification by providing scores for individual attributes. 

\textbf{Vision-Language Models.} Large-scale vision-language models (VLMs), such as CLIP~\cite{radford2021learning}, have bridged the visual and language modalities through contrastive learning.
VLMs estimate the similarity between text and an image, leading to many downstream applications, such as zero-shot classification with open-ended language.
Several improvements have been proposed to the VLMs, such as training with noisy data~\cite{jia2021scaling,cherti2023reproducible,tsimpoukelli2021multimodal}, better training strategies~\cite{alayrac2022flamingo,wang2021simvlm,desai2021virtex,chen2022visualgpt,li2019visualbert,xu2023bridgetower,tan2019lxmert,lu2019vilbert}, concept localization~\cite{zeng2021multi}, multiple modalities~\cite{yuan2021florence}, grounding abilities~\cite{li2022grounded,li2024desco}, or generative capabilities~\cite{li2022blip,li2023blip,mokady2021clipcap,luo2022frustratingly}. 
Another advantage of VLMs is that classification or retrieval can be done by using a list of attribute descriptors in conjunction with the class names~\cite{menon2022visual, pratt2023does, yan2023learning}. 
Classification by using descriptions results in an intermediate step of interpretability, as these lists of descriptions are essentially concept bottlenecks.
Waffle-CLIP~\cite{roth2023waffling} proposed using random words along with these concepts leads to more robust concept embedding.
However, all these methods have access to a knowledge base (large language models in this case), which they use to retrieve descriptions for categories using the category name.
Such approaches fail if the categories are never discussed or are infrequently mentioned on the internet.
Since our approach does not rely on the name of the category, we perform better on specialized and esoteric categories.

\textbf{LLMs as Optimizers.} Using a large language model to generate mutations with evolutionary search has been implemented for the task of code generation \cite{romera2023mathematical}. We follow a similar approach, except for the task of visual recognition. In addition, their fitness functions are a series of non-learning-based metrics. We instead use another foundation model, a vision-language model, as the fitness function, thereby scaling to open-ended visual concepts. 
Another instance in which LLM's are used as optimizers is for prompt discovery \cite{yu2023language}. Similarly, concurrent work leverages LLMs as optimizers to find attributes to improve visual classification, but crucially, they provide the class name as an input to the large language model \cite{jin2024llms,han2023llms}. As such, they use privileged information that we do not assume. 

\section{Discovering Visual Classifiers}

\begin{figure}[t]
  \centering
  \includegraphics[width=1.0\textwidth]{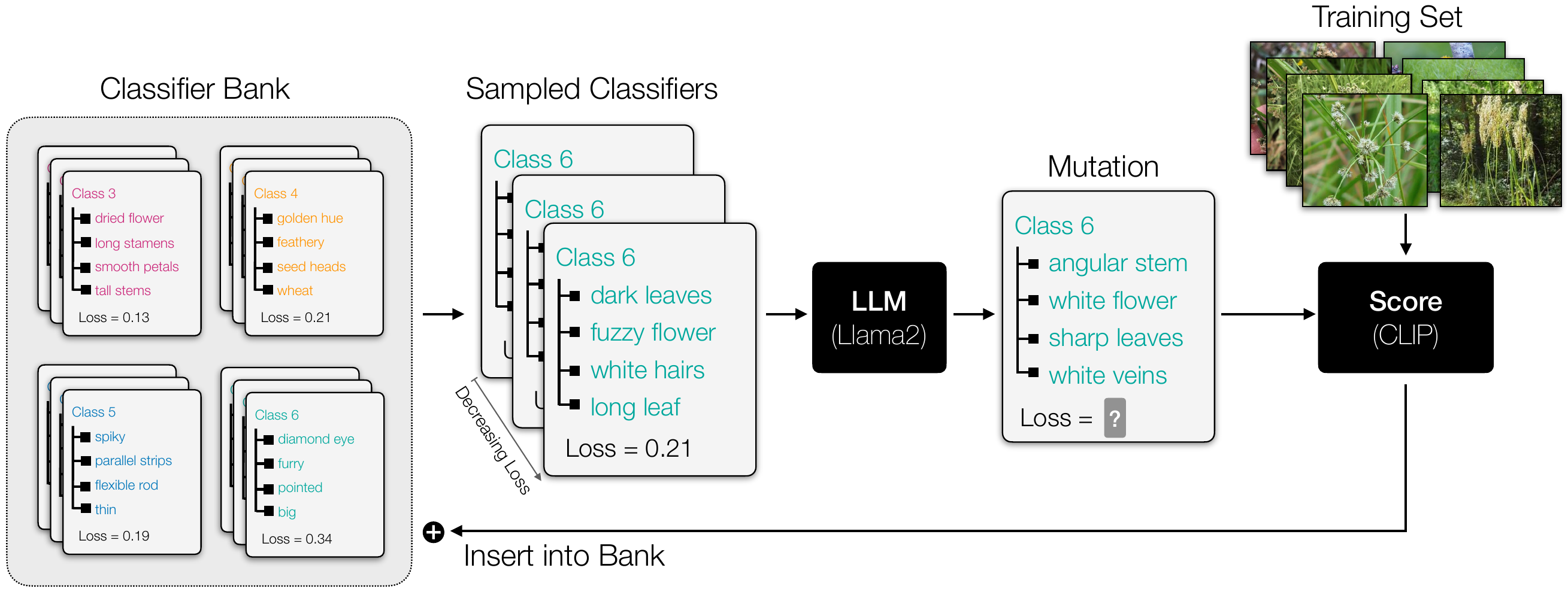}
  \caption{\textbf{Method.} LLM-Mutate is an evolutionary algorithm that learns sets of discrete language attributes per class. The mutation and cross-over operations, which are mechanisms to introduce new parameter hypotheses, are replaced by a large language model that uses in-context learning over past attributes and their scores to iteratively generate better attributes. }
  \label{fig:method}
\end{figure}

\subsection{Model}

Given an image $x$, our goal is to predict its category label $y$. We create a concept bottleneck model $f_c(x)$ that scores whether the image $x$ contains category $c$. Once optimized, the concept bottleneck model should produce a high score for $c = y$ and a low score otherwise. To score the class $c$ from the image $x$, we average over a bottleneck of discrete attributes:
\begin{align}
    f_c(x; \mathcal{D}) =  \frac{1}{\left| \mathcal{D}(c) \right|} \sum_{d_i \in \mathcal{D}(c)} \phi(d_i, x)
    \label{eqn:model}
\end{align}
where $\mathcal{D}(c)$ is the set of attributes to recognize class $c$, and $\phi(d_i, x)$ is the score from a vision-language model (such as CLIP) to detect the attribute $d_i$ in image $x$. The model in \cref{eqn:model} provides a degree of explainability because a prediction must be based in natural language attributes $d_i\in \mathcal{D}(c)$.
At inference, we perform multi-class classification by scoring \cref{eqn:model} for each class, and picking the highest scoring one, i.e.\ $\arg\max_c f_c(x; \mathcal{D})$.

Concept bottleneck models have traditionally been challenging to implement in computer vision because we need to instantiate a discriminative set of discrete attributes $\mathcal{D}(c)$, which pose challenges for gradient-based optimization methods that are now ubiquitous in deep learning.  Prior work has relied on manual annotation of $\mathcal{D}(c)$ for each class \cite{koh20concept}, which does not efficiently scale, or relied on other knowledge bases \cite{menon2022visual}, which cannot generalize to specialized categories. 

\subsection{Learning and Optimization}

Given $C$ categories (without semantic labels), we want to learn discriminative attributes $\mathcal{D}$ where $\mathcal{D}(c)$ are the attributes for class $c$.  We optimize the objective:
\begin{align}
    \min_{\mathcal{D}} \; \mathbb{E}_{(x, y)} \left[
    \mathcal{L}\left( \hat{y}, y \right)
    \right] \quad \textrm{for} \quad \hat{y} = \left[ f_1(x; \mathcal{D}), \ldots, f_C(x; \mathcal{D}) \right],
    \label{eqn:objective}
\end{align}
where $\mathcal{L}$ is a loss function that measures the error of the predictions $\hat{y}$ to the label $y$ of each training example. We use the cross-entropy loss for $\mathcal{L}$. 

Since attributes for each class are discrete and $f_c$ is not differentiable, we optimize \cref{eqn:objective} with evolutionary search. We maintain a bank of hypotheses $B$ for potential $\mathcal{D}(c)$, mutate them, and keep the best-scoring hypotheses according to the objective function. Typically, evolutionary search creates heuristics to mutate the attributes $\mathcal{D}(c) \in B$, for example by randomly merging attributes together (called crossover) or by randomly injecting new words from a vocabulary. However, these heuristics are not efficient for two reasons. First, the search space of natural language descriptions is large. Second, most heuristics do not leverage patterns between attributes and their performance that could drive the optimization to convergence rapidly.

We propose to replace the mutation step with a large language model and its in-context learning capabilities. Given $k$ past hypotheses $\{ \mathcal{D}_t(c), \ldots, \mathcal{D}_{t-k}(c) \}$ for a class $c$, and their loss, we ``mutate'' the next hypotheses through:
\begin{align}
    \hat{\mathcal{D}}_{t+1}(c) = \textrm{LLM}\left(\mathcal{D}_t(c), \ldots, \mathcal{D}_{t-k}(c)\right)
\end{align}
where $\mathcal{D}_t$ is an attribute set from iteration $t$. By using an LLM, the mutated $\hat{\mathcal{D}}_{t+1}(c)$ benefit from natural language priors, allowing us to efficiently search for descriptors that obey the semantics and syntax of natural language. Secondly, the in-context learning ability of the LLM means they will be able to find patterns in the past hypothesis to guide the search towards attributes that are likely to minimize the objective function. After mutation, we add $\hat{\mathcal{D}}_{t+1}(c)$ to the bank of classifiers if it improves the objective and iterate.

We use the open-source Llama-2-70B-Instruct \cite{touvron2023llama} for the large language model $\textrm{LLM}(\cdot)$ and CLIP ViT-B/32 \cite{radford2021learning} as our vision-language model $\phi(\cdot)$. The starting attributes are initialized randomly, with no prior knowledge of the class name or prior information.

\begin{algorithm}[t!]
\caption{Discriminative Attribute Set Learning}\label{alg:three}

\DontPrintSemicolon 

\KwIn{Number of categories $C$, Loss function that scores attribute set $\mathcal{L}(\mathcal{D})$ on the training set, Scalar hyper-parameters $N$, $M$} 
\KwOut{Discriminative attributes $\mathcal{D}_{\textrm{best}}$, where $\mathcal{D}_{\textrm{best}}(c)$ is the set of attributes to recognize class $c$} 

\tcp{Randomly initialize the classifier bank $B$} 
$B \gets \{\}$ \;
\For{$i = 1,\ldots,N$}{
    \For{$c = 1,\ldots,C$}{
        $\mathcal{D}_i(y) \leftarrow \textrm{randomly sampled attributes} $ \;
    }
    $B \leftarrow B \; \cup \; \{ \mathcal{D}_i \}$ 
} 

\tcp{Learn a $\mathcal{D}$ that discriminates the training set $\{(x, y)\}$} 

\While{\KwSty{not converged}}{
    \tcp{Biased sampling of $M$ classifiers by expected loss $\mathbb{E}\left[\mathcal{L}\right]$}
    $S = \{\mathcal{D}_{1}, ... , \mathcal{D}_{M}\}$ where $\mathcal{D}_i \sim B \;\; \textrm{s.t.} \;\; p(\mathcal{D}_i) \propto \mathbb{E}\left[\mathcal{L}(\mathcal{D}_i)\right]$\; 
    \tcp{Sort by the expected loss}
    $S =  (\mathcal{D}_{1}, ..., \mathcal{D}_{M})$ where $\mathbb{E}\left[\mathcal{L}(\mathcal{D}_i)\right] \geq \mathbb{E}\left[\mathcal{L}(\mathcal{D}_j)\right] \;\; \forall_{i<j}$\; 
    \tcp{One step of evolutionary search for each category}
    \For{$c = 1,\ldots,C$}{
        \tcp{Mutate the attributes with the LLM}
        $S_c = (\mathcal{D}_{i}(c) : \mathcal{D}_{i} \in S)$\; 
        $\hat{\mathcal{D}}(c) \gets \textrm{LLM}(S_c)$\;

        $S' = (\mathcal{D}'_1, \ldots, \mathcal{D}'_M)$ where $\forall_k \; \mathcal{D}'_i(k) = \begin{cases}
            \hat{\mathcal{D}}(c) & \textrm{if } c = k\\
            \mathcal{D}(k) & \textrm{otherwise}
        \end{cases}$

        \tcp{Keep best attribute set}
        $B \gets B \; \cup \; \arg\min_{\mathcal{D} \in S'} \{\mathbb{E}\left[\mathcal{L}(\mathcal{D})\right] \}$\;
    }
}

\Return $\arg\min_{\mathcal{D} \in B} \{\mathbb{E}\left[\mathcal{L}(\mathcal{D})\right] \}$\;

\end{algorithm}







\subsection{Evolutionary Search}

\cref{alg:three} shows the evolutionary search procedure to optimize the attribute sets for image classification.  We first initialize the classifier bank $B$ with random words to create $N$ initial hypotheses for $\mathcal{D}_i$.
During learning, we sample sets of attributes from $B$ to construct the in-context examples for the mutation step, and repeat this process until convergence. We bias the samples according to the loss of each attribute set (normalized with the softmax operation).
For each sampled set, we construct a prompt per class by concatenating the attributes from class $c$ in increasing order of performance. We prompt the LLM separately per class, which generates the novel, mutated attributes, $\hat{\mathcal{D}}(c)$ for each class.
We evaluate the newly mutated attribute sets and add the mutated classifier with the lowest loss to the classifier bank. We provide more implementation details in the \href{https://llm-mutate.cs.columbia.edu/static/pdfs/supplementary.pdf}{supplementary material}.

\begin{figure}[b!]
  \centering
  \includegraphics[width=1.0\textwidth]{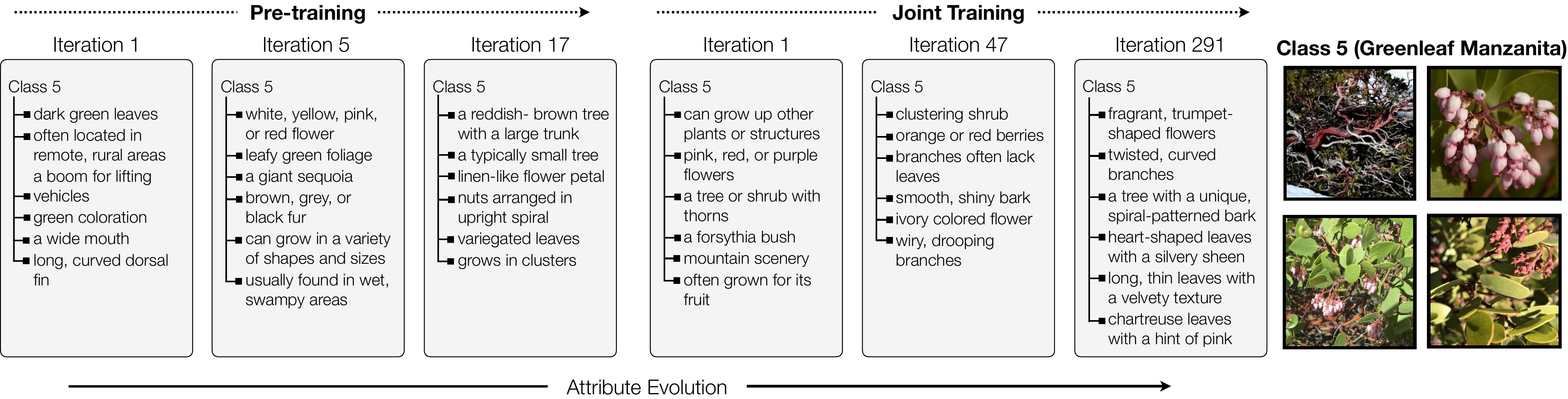}
  \caption{\textbf{Attribute evolution.} We show examples of the attribute evolution for both the pre-training and joint-training stages of learning. At the beginning, the first generated set of attributes have little to do with the class, and by the end of the joint-training, the learned attributes are  specific to the Greenleaf Manzanita.}
  \label{fig:attribute-evolution}
\end{figure}

\subsection{Classifier Bank Initialization}

During fine-grained classification, the optimization favors attributes that are highly discriminative between visually similar classes, causing attributes common to multiple classes to be discouraged. We want to encourage common attributes at the beginning of learning so that the optimization discovers class-specific details, instead of spurious unrelated attributes arising from noise.

To achieve this, we use a pre-training strategy to first discover common attributes, which serves as the initialization for joint multi-class training. 
We implement the pre-training step by learning a binary classifier per class, using the same evolutionary algorithm as before, but with an objective function to separate one class from all others, including significantly unrelated classes. We use the following objective for pre-training: 
\begin{align}
    \min_{\mathcal{D}} \; \mathbb{E}_{x_{p}} \left[
    f_{c}\left( x_{n}; \mathcal{D} \right)\right] - \mathbb{E}_{x_{n}} \left[
    f_{c}\left( x_{p}; \mathcal{D}\right)
    \right] 
    \label{eqn:pre-train}
\end{align}
where $x_p$ is a positive images of the class $c$, and $x_n$ denotes the negative images of all the other classes. Crucially, we include classes outside of the fine-grained dataset in the negative classes here as well.
The generated attributes across the first $200$ iterations become the initialization of the set of attributes for the program bank initialization of the joint training. We randomly initialize the attribute bank for binary-pre-training with a large pool of attributes generated by an LLM about generic visual categories. \cref{fig:attribute-evolution} shows an example of the initialization, and we include further implementation details in the supplementary.


\begin{figure}[t!]
  \centering
  \begin{minipage}[t]{0.95\textwidth}
    \centering
    \includegraphics[width=\linewidth]{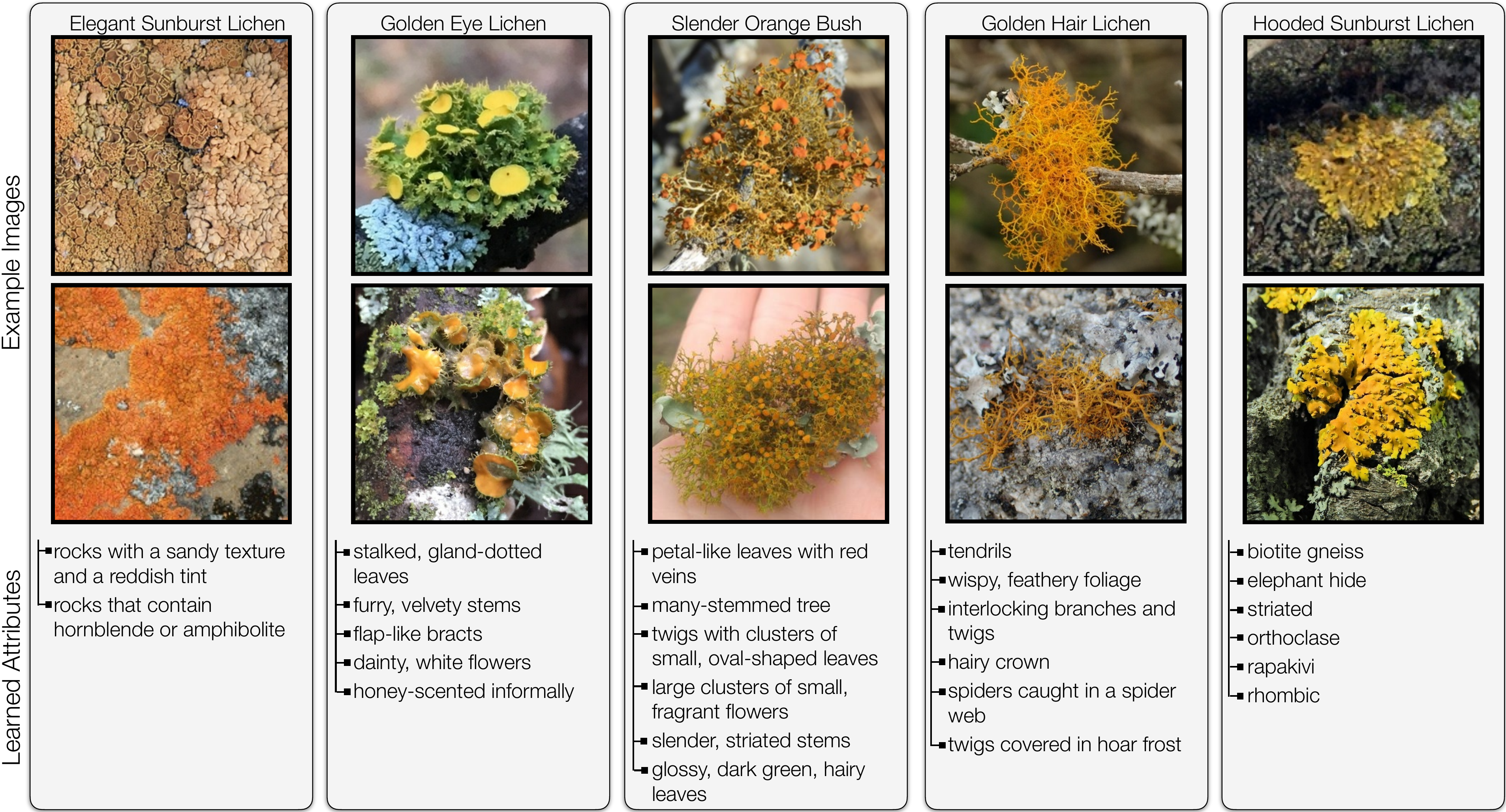}
  \end{minipage}
  
  \begin{minipage}[t]{0.95\textwidth}
    \centering
    \includegraphics[width=\linewidth]{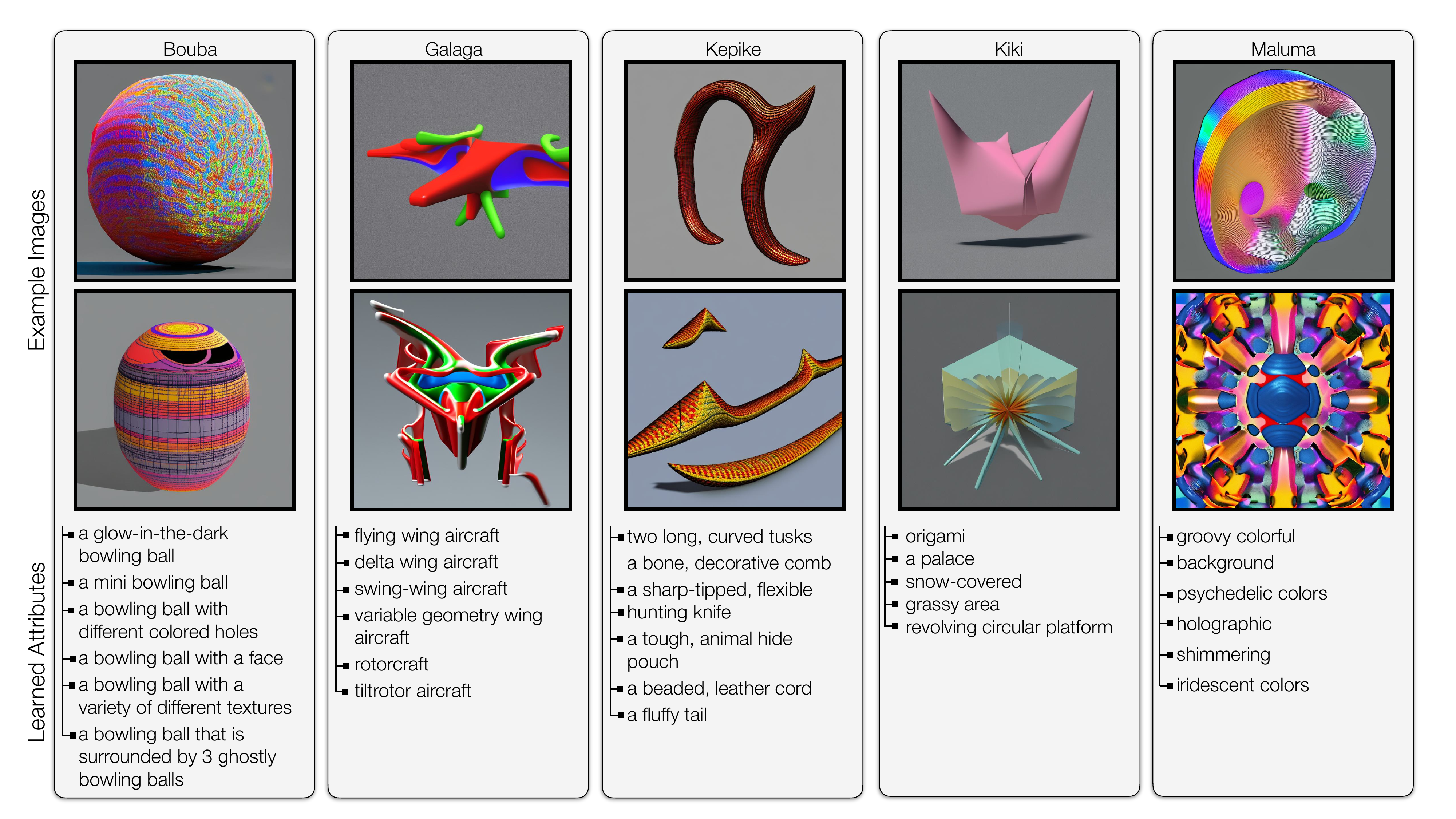}
  \end{minipage}
  \caption{\textbf{Qualitative Results.} We show qualitative results for the iNaturalist Lichen family and a KikiBouba dataset. The results illustrate two sample images per class and the learned attributes. The learned attributes for the Lichen hardly refer to color, as this is a common feature to all Lichen, and instead focus on structural properties.}
  \label{fig:qualinat}
\end{figure}

\section{Experiments}

\subsection{Datasets}
We validate our method by evaluating on two different datasets. 
First, we evaluate on subsets of iNaturalist -- a fine-grained classification dataset with rare species that are rarely discussed online.
We then evaluate on an image dataset of novel concepts, created with new words that are even less likely to be discussed online, as they are invented.

\textbf{iNaturalist:} 
iNaturalist~\cite{van2021benchmarking} is a dataset for fine-grained species classification.
It contains images and annotations obtained from citizen scientists for a large number of animal, plant, and fungus species.
We experiment with five different families and classify between five to six species with each family. We chose families whose features varied in more complex ways than color.

\under{Lichen (fungi)} has 6 species: elegant sunburst, golden-eye, slender orange-bush, golden-hair, hooded sunburst, and maritime sunburst lichen.
\under{Wrasse (fish)} has 5 species: caribbean bluehead, six-bar, cortez-rainbow, moon, and ornate wrasse.
\under{Wild rye (grass)} has 5 species: squirreltail, bottlebrush, quack grass, canada wild rye, and virginia wild rye.
\under{Manzanita (berry shrubs)} has 5 species: big-berry, pinemat, greenleaf, point-leaf, and pine-mat manzanita.
\under{Bulrush (herb)} also has has 5 species: dark-green, woolgrass, panicled, rufous, and wood bulrush.

\textbf{Kiki-Bouba:} 
We also validate our method on completely novel concepts that do not appear on the internet and to which language models lack familiarity. In a surprising study, the Kiki-Bouba experiment \cite{ramachandran2001synaesthesia} showed that people tend to associate specific symbols with different sounds, even though the words and the physical objects do not exist. We create a dataset of images corresponding to non-existent concepts with generative models trained to generate images from meaningless words ~\cite{alper2024kiki}. 
Such a dataset with novel concepts makes it a strong testbed for attribute discovery.

\subsection{Baselines}

We compare our method against several baselines. Every baseline that is not our own has access to privileged information that we do not. Specifically, the baselines with zero-shot attributes were generated by prompting GPT3 \cite{mann2020language} with the class name. The baselines that contain class names have an evident advantage. Lastly, the gradient-based approach we constructed does not have access to privileged information such as class names. Nonetheless, our evolutionary method significantly outperforms this baseline for every dataset, thus justifying our evolutionary approach. Our results show that our method outperforms all of the baselines across all datasets, demonstrating that it can learn attributes for specialized and undocumented visual categories.

\textbf{Class Name (CLIP \cite{radford2021learning}):} Our first baseline is the simple method of classifying with the class name. For the iNaturalist dataset, which has both a common and a scientific name per species, we report the best accuracy between using the common name, the scientific name, and both names.

\textbf{Classification by Description \cite{menon2022visual}:} We implement the Classification by Description method, proposed by \cite{menon2022visual}, which generates zero-shot visual attributes for a class by prompting GPT3 \cite{mann2020language}, and joins the class name to each of the zero-shot attributes. Similarily to the ``Class Name'' baseline, for the iNaturalist dataset, we report the best accuracy between using the common name, the scientific name, and both names. 

\textbf{Zero-shot Attributes:} We also use a variant of Classification by Description where the class name is no longer appended to the attributes.

\textbf{Gradient-based Approach:} Instead of using an LLM to search for attributes, we instead search for optimal input tokens to the text encoder of the VLM using gradient descent. 
For a class, we find tokens that highly discriminate it from other classes. 
The model optimizes for probability values over the complete token list. 
However, since the explanations have to be tokens and not a probability distribution over them, we enforce the probability distribution to be more selective to fewer tokens as training progresses by using a temperature parameter that decreases over the training period.

\begin{table}[t]
  \caption{We report the accuracy per dataset, for our method and baselines.
  }
  \label{tab:table1}
  \centering
  \begin{tabular}{@{}llllllll@{}}
    \toprule
    & \multicolumn{5}{c}{iNaturalist} & \multicolumn{2}{c}{Kiki-Bouba} \\
    \cmidrule(lr){2-6} \cmidrule(lr){7-8}
    Method & Lichen & Wrasse & Wild Rye & Manzanita & Bulrush & KB1 & KB2 \\
    \midrule
    Zero-shot Attributes & 28.3 & 16.0 & 22.0 & 18.0 & 24.0 & 20.6 & 19.2\\
    Class Name (CLIP \cite{radford2021learning})& 23.3 & 32.0 & 32.0 & 26.0 & 26.0 & 38.7 & 38.8 \\
    Classification by Desc. \cite{menon2022visual} & 30.0 & 34.0 & 36.0 & 28.0 & 20.0 & 28.8 & 36.8\\ 
    Gradient-based Approach & 23.3 & 20.0 & 40.0 & 20.0 & 20.0  & 16.7 & 55.6\\
    Ours (1-prompt) & 31.6 & 24.0 & 44.0 & 40.0 & 22.0 & 50.3 & 47.8\\
    Ours (10-prompt) & \textbf{48.3} & \textbf{44.0} & \textbf{58.0} & \textbf{58.0} & \textbf{42.0} & \textbf{79.2} & \textbf{59.4}\\
  \bottomrule
  \end{tabular}
\end{table}

\begin{figure}[tp]
  \centering
  \includegraphics[width=1.0\textwidth]{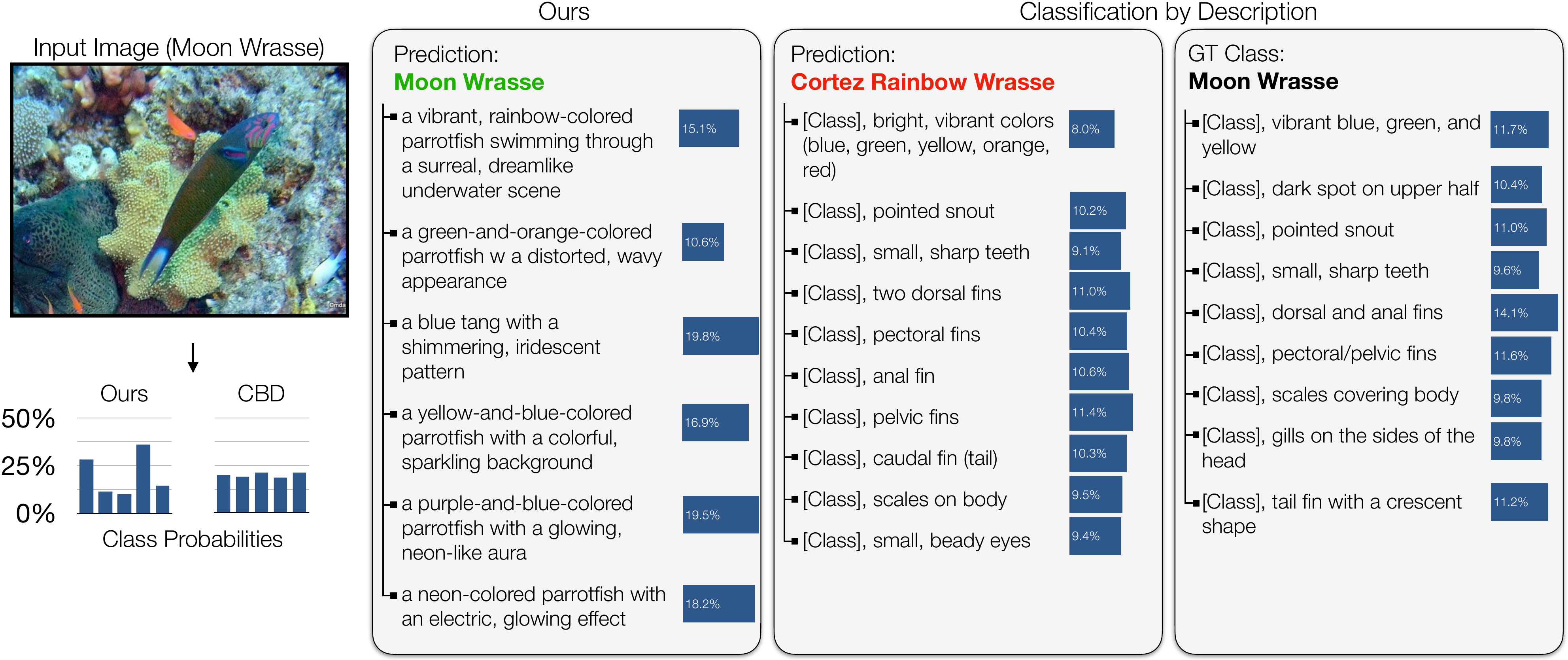}
    \includegraphics[width=1.0\textwidth]{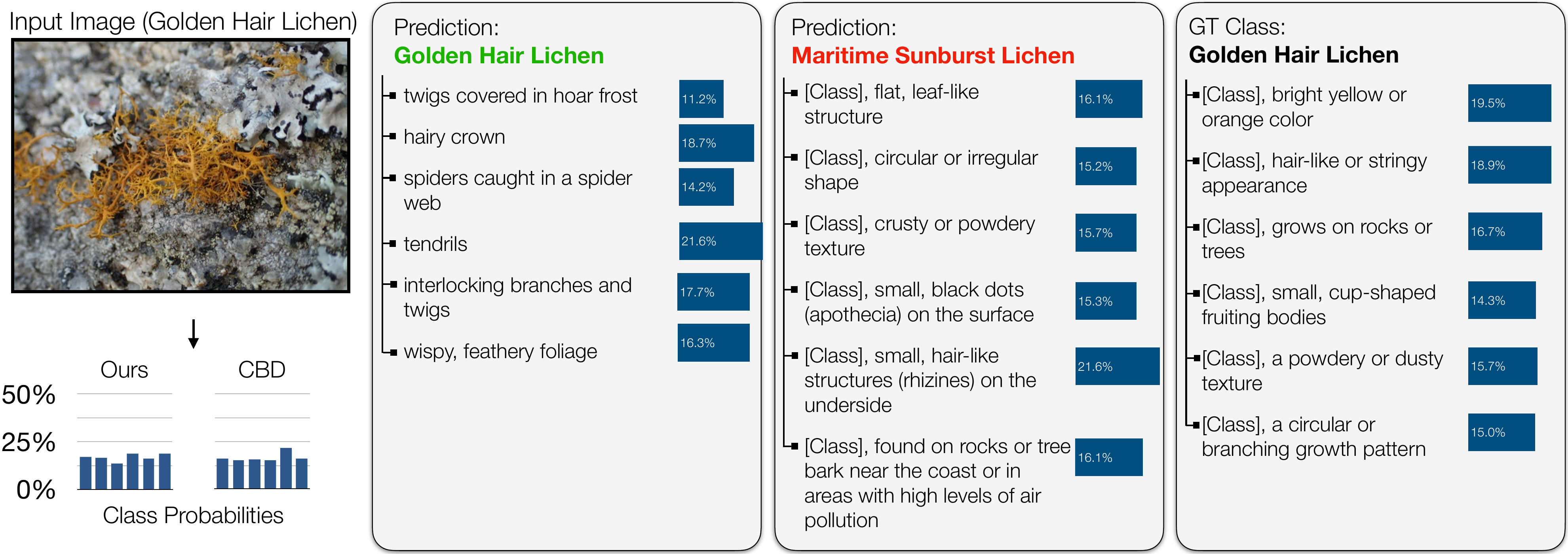}
\includegraphics[width=1.0\textwidth]{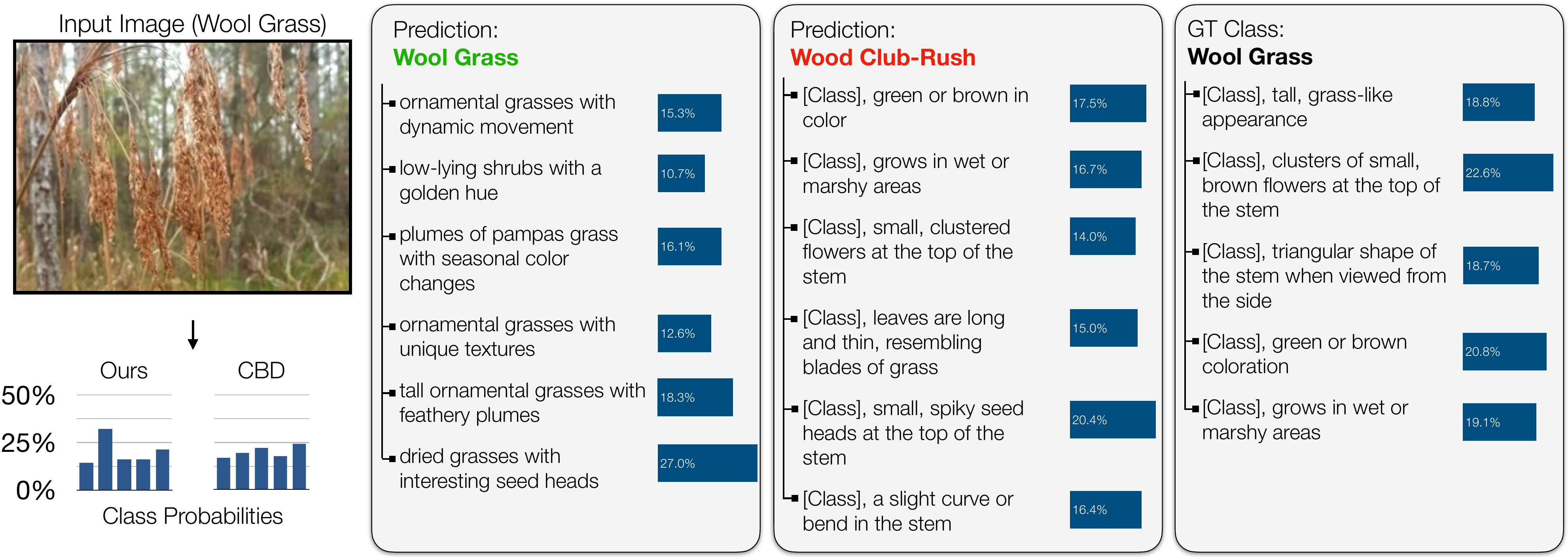}
  \caption{\textbf{Predictions.} We show three different prediction examples. For each example, we show our method's prediction (first column), as well as classification by description \cite{menon2022visual} (CBD)'s prediction (second column), and CBD's attributes for the ground truth class (third column). For each column, we show the normalized probability per attribute. Below the input image, we show the probability distributions across classes for both our method and CBD. The results show that our learned attributes are more detailed and discriminative of the species within the family, compared to the description by classification (CBD) baseline. Furthermore, our method's class probability distributions tend to be more concentrated than CBD's. }
  \label{fig:preds-figure}
\end{figure}

\textbf{Varying Prompt Length:} We report the results of our method with two different prompt lengths: one and ten. In the former, the LLM prompt only has one example of a set of attributes for a particular class, and therefore doesn't see the increasingly better sets of attributes. We notice a significant drop in accuracy when in-context learning is prohibited. 

\textbf{Engineered Text Templates:}
For each method, we report the best accuracy per dataset between the accuracy computed using scores averaged over the engineered text templates proposed in CLIP \cite{radford2021learning}, and without averaging over the engineered text templates. We outperform all baselines with or without engineered text templates, and report the full numbers in the supplementary.

\begin{figure}[b!]
  \centering
  \includegraphics[width=1.0\textwidth]{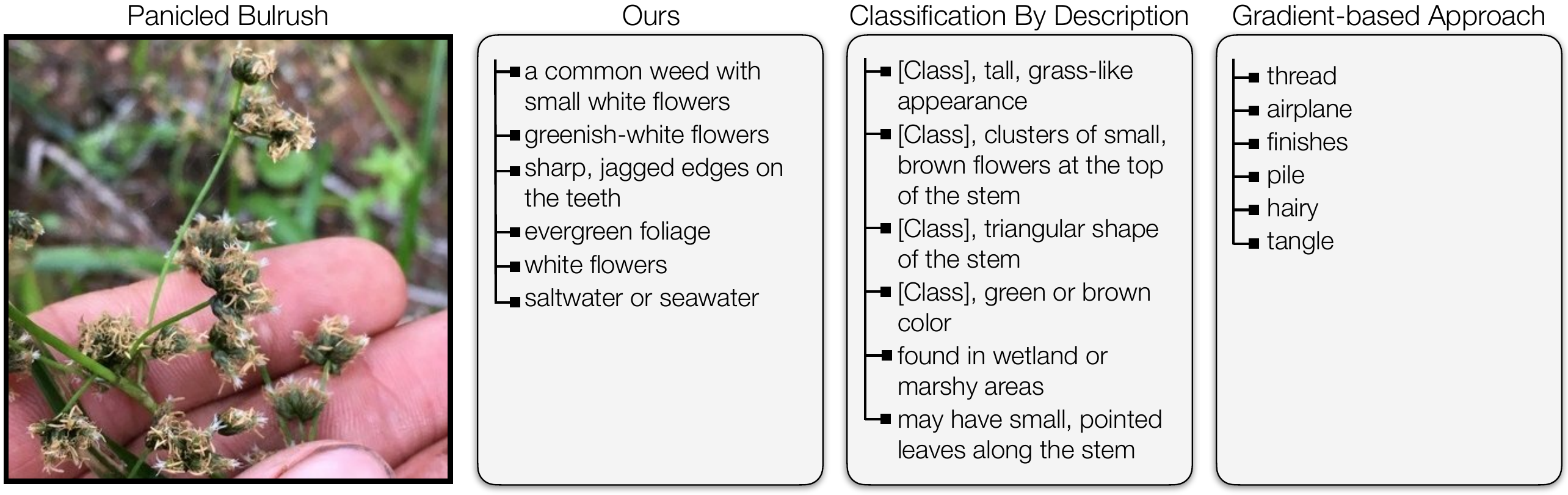}
  \caption{\textbf{Comparison of Attributes by Method}. We show qualitative examples of our learned attributes, classification by description's attributes (CBD), and our gradient-based approach attributes. CBD often produces reasonable attributes, but they are not discriminative, resulting in poor recognition accuracy. Gradient-based methods often produce poor attributes due to optimization difficulties.}
  \label{fig:threemethods}
\end{figure}

\subsection{Fine-Grained Classification (iNaturalist)}

\textbf{Quantitative Result Discussion:} We present the performance on accuracy for our method and baselines in \cref{tab:table1}. Crucially, our method starts out with zero prior knowledge on any information as to what is in a class of images, beyond the fact that those images are grouped together. The initialized best classifier at the beginning of optimization has no correspondence to the class name or quality, and it is only through optimization that our method learns to discover interpretable and reflective attributes. The first three baselines of zero-shot descriptors, class name, and class name with zero-shot descriptors on the other hand have prior knowledge on the class from the very start, and therefore are at a significant advantage as they wouldn't work for any concepts that the internet isn't already familiar with. \newline 
\newline 
 \noindent \textbf{Qualitative Result Discussion:} In \cref{fig:qualinat}, we show the fine-grained classification results on the Lichen family. We observe that while all the Lichen are orange and yellow in color, the only references to color within the attributes are when the color diverges from the mean color, i.e. ``reddish tint''. The majority of the attributes reflect the Lichen's structure, which is due to the fact that all the Lichen are roughly the same color and grow in similar environments, the principal discriminating feature is the structure. In \cref{fig:threemethods}, we compare our learned attributes to the zero-shot attribute in CBD, as well as the learned attributes with our gradient-based approach. The gradient-based approach has far less descriptive and interpretable attributes. We provide more such qualitative examples of the three methods in the supplementary material.

In \cref{fig:preds-figure}, we explicitly visualize the predictions across for three of the classifiers, one per row. Each example illustrates how each method compares to the top-performing baseline, classification by description (CBD). We show our method's prediction as well as CBD's prediction, along with the relative contribution of each attribute to the mean score. We additionally denote CBD's attributes for the ground truth class, along with the relative contribution to the ground truth mean score, to visualize why CBD may have incorrectly predicted the class. Across the three examples, we notice that CBD's attributes are less specific and detailed compared to ours, and that there are many shared attributes across different classes. Shared attributes are not useful in fine-grained classification, since the goal is to discriminate between classes. 
 
\subsection{Novel Objects (Kiki-Bouba)}

\textbf{Quantitative Result Discussion:} We outperform all baselines on the two KikiBouba datasets. We notice that the gradient-based baseline has a large variation in performance, and we hypothesize that this is due to the intra-class variation changing quite significantly across classes. Across all experiments, the gradient-based approach produces poor attributes due to optimization difficulties, as illustrated in \cref{tab:table1}.\newline 
\newline 
\textbf{Qualitative Result Discussion:} For the novel image results on the first dataset of Kiki-Bouba. In \cref{fig:qualinat}, we show qualitative results of the learned attribute sets for the second Kiki-Bouba dataset. We notice that compared to the iNaturalist datasets, there is less similarity between classes, and the discovered attributes are more object-oriented. We suspect these observations are not unrelated, and that when an object name can be used to discriminate between classes, the vision-language model scores highly with it.

\begin{figure}[t!]
    \begin{minipage}{0.65\textwidth}
        \centering
        \includegraphics[width=1.0\textwidth]{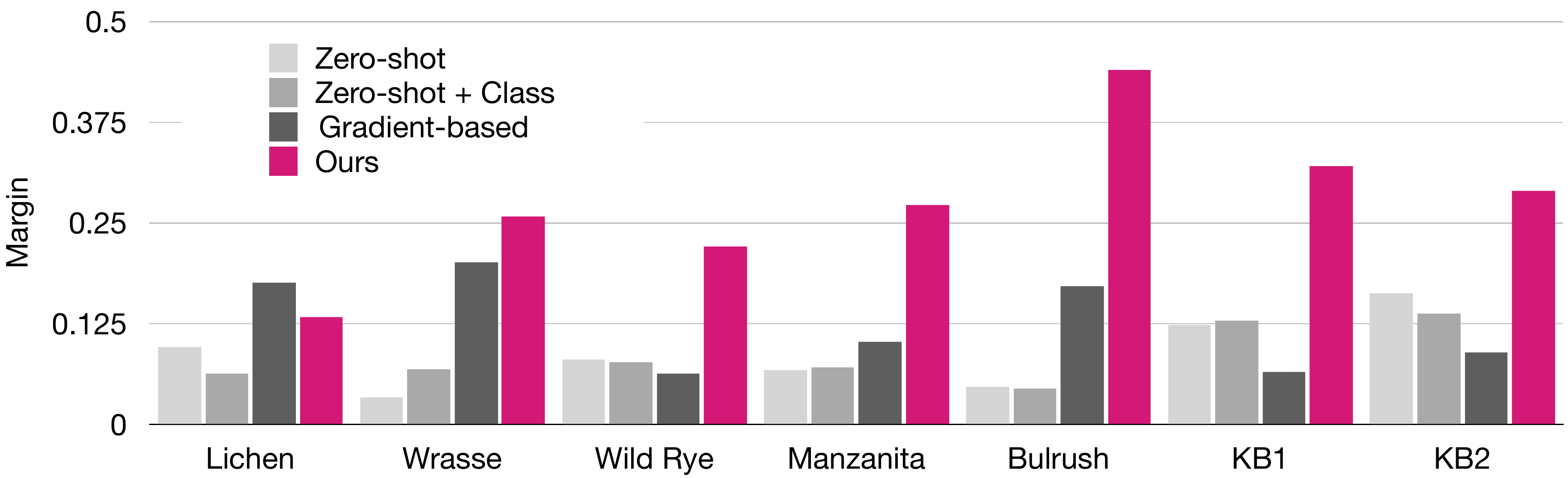}
    \end{minipage}\hfill
    \begin{minipage}{0.34\textwidth}
        \caption{\textbf{Confidence}. We show the mean margin of the scores between the top and runner-up prediction, which measures the typical confidence of each model.}
        \label{fig:margin}
    \end{minipage}
\end{figure}
\subsection{Analysis of Learning}
\textbf{Margin Metric:}  In \cref{fig:preds-figure}, below the input image, we show the probability distributions across classes for both our method and CBD. We observe that in general, the probability distribution across classes for CBD is more uniform than ours. We investigate this observation by measuring the \emph{margin} metric per dataset, per method. The margin is defined as $\max_{c \in C} f_c(x_i; \mathcal{D}) - \max_{\bar{c} \notin C} f_{\bar{c}}(x_i; \mathcal{D})$, which measures the difference in scores between the top prediction and the runner-up, indicating the prediction confidence \cite{melville2004diverse}. We plot the margin for each method, per dataset, in \cref{fig:margin}. The results illustrate that our method is on average twice as confident as the other methods. \newline

\begin{figure}[t!]
  \centering
  \includegraphics[width=1.0\textwidth]{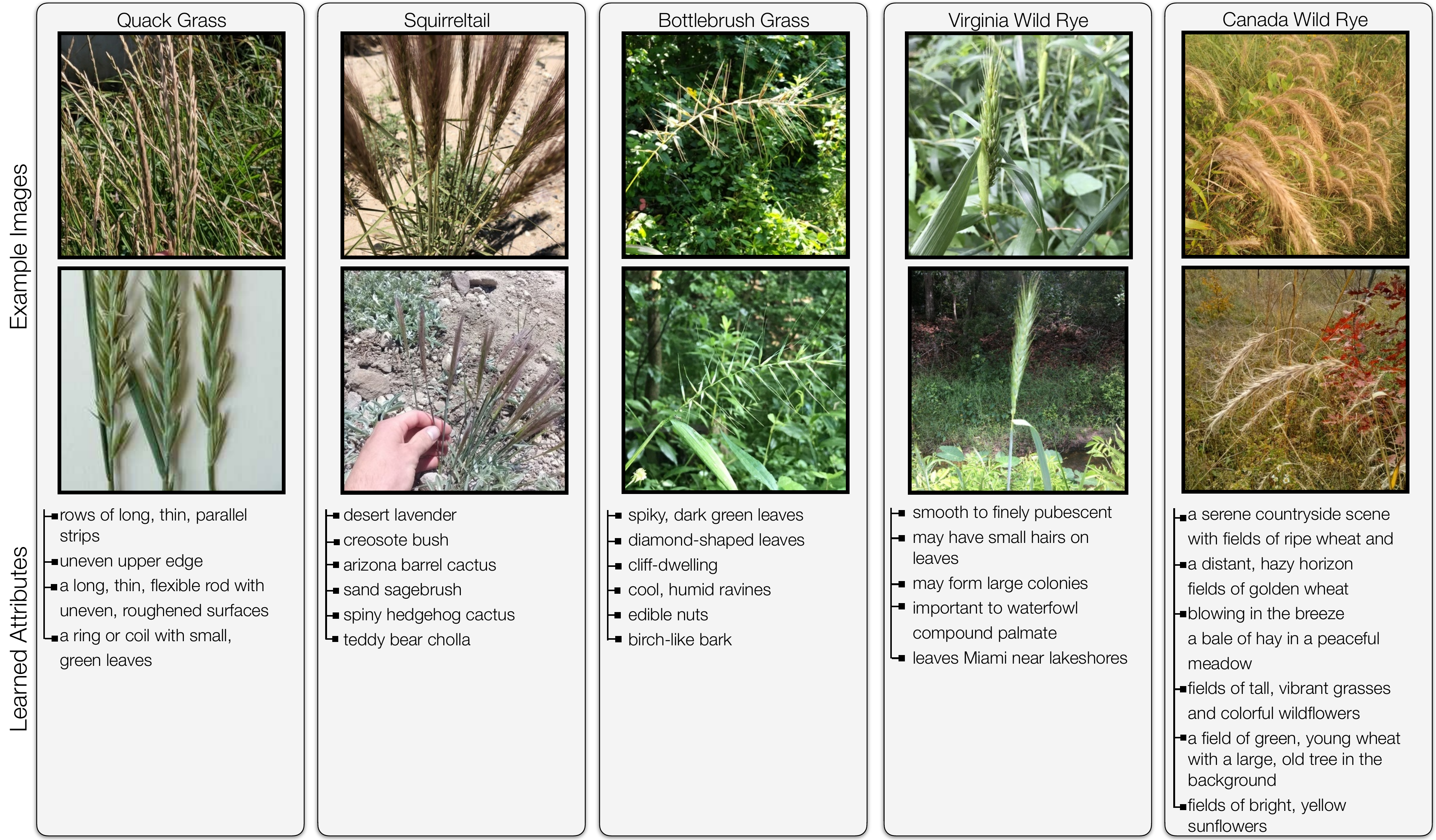}
  \caption{\textbf{Dataset Bias}. The \textbf{squirreltail} species is the only species that commonly lives in drought habitats amongst the family, and the learned  attributes are names of plants that live in the desert. The ability to explicitly audit bias is an advantage of our interpretable method.}
  \label{fig:datasetbias}
\end{figure} 

\noindent\textbf{Attribute Evolution:} In \cref{fig:attribute-evolution}, we show examples of the iterations for pre-training, followed by the joint training. At initialization, the first sampled set of attributes for Greenleaf Manzanita hardly relate to the class. The closest attributes are ``dark green leaves'' and ``green coloration'', with the other attributes having nothing in common with images. However, by the end of the attribute evolution, the color ``green'' is no longer part of the attributes, as the other Manzanita also have green leaves, making it non-discriminative. The final attributes contain descriptors that are particularly descriptive of the Greenleaf Manzanita. We share more examples of attribute evolution in the supplementary.
\newline

\subsection{Auditing Dataset Bias}

By having explicitly interpretable attributes as the bottleneck for classification, we can directly observe whether the classifier is picking up on dataset bias to perform prediction. An example of dataset bias can be seen in \cref{fig:datasetbias}. This is unique to our method, as classifiers with no concept bottlenecks have no way of converting dataset bias into language, and previous work in concept bottlenecks do not discover the attributes from the data itself. 

\section{Discussion}

\textbf{Societal Impacts and Limitations.} Explainable vision systems have the potential to have significant practical impact, especially in specialized, critical, and scientific domains.  Such interpretable classifiers can establish trust, as they provide insight into how a classifier reached a decision. They allow people to audit the decision-making process, which is important for many practical cases. Finally, they also impact education, as the classifiers can report to a person the visual differences it has discovered, thereby helping the person learn about the recognized concept too. However, since our approach uses open-source LLMs, our method inherits known limitations about LLMs in bias and inappropriate generations.  As research in LLMs advance, we expect our framework to improve too. \newline
\newline 
\noindent\textbf{Conclusion.} We propose a framework that integrates large language models and evolutionary search in order to learn interpretable, discrete attributes for visual recognition. In multiple datasets, our method outperforms existing baselines significantly.


%
%
\bibliographystyle{ieeetr}
\bibliography{main}
\end{document}